\documentclass[]{article}

\usepackage{times}
\usepackage{graphicx}
\usepackage{url}
\usepackage{amsfonts}
\usepackage{moreverb}
\usepackage{amsmath}
\usepackage{caption}
\usepackage{bm}
\usepackage{bbm}
\usepackage{paralist}
\usepackage{subfig}
\usepackage{array}
\usepackage{booktabs}
\usepackage[table]{xcolor}
\usepackage[ruled]{algorithm2e}
\usepackage{dsfont}
\usepackage{xcolor}
\usepackage{booktabs}
\usepackage{float}
\usepackage{authblk}
\usepackage{multicol}
\usepackage[margin=1in]{geometry}

\newcolumntype{L}{>{$}l<{$}}
\newcolumntype{C}{>{$}c<{$}}
\newcolumntype{R}{>{$}r<{$}}

\renewcommand\vec{\mathbf}
\DeclareMathOperator{\E}{\mathbb{E}}
\newcommand{\R}{\mathbb{R}}
\DeclareMathOperator{\argmax}{arg\,max} % Jan Hlavacek

\def\stoptable#1{%
	\par\vspace{\abovecaptionskip}%
	{\footnotesize #1}%
}

\title{A Perturbation Resistant Transformation and Classification System for Deep Neural Networks}

\author[1]{Nathaniel Dean}
\author[2]{Dilip Sarkar}
\affil[1,2]{Department of Computer Science, University of Miami}
\affil[1]{\textit{nxd551@miami.edu}}
\affil[2]{\textit{sarkar@cs.miami.edu}}

\begin{document}
\date{}
\maketitle

\begin{abstract}
Deep convolutional neural networks accurately classify a diverse range of natural images, but may be easily deceived when designed, imperceptible perturbations are embedded in the images.  In this paper, we design a multi-pronged training, input transformation, and image ensemble system that is attack agnostic and not easily estimated.  Our system incorporates two novel features.  The first is a transformation layer that computes feature level polynomial kernels from class-level training data samples and iteratively updates input image copies at inference time based on their feature kernel differences to create an ensemble of transformed inputs.  The second is a classification system that incorporates the prediction of the undefended network with a hard vote on the ensemble of filtered images.  Our evaluations on the CIFAR10 dataset show our system improves the robustness of an undefended network against a variety of bounded and unbounded white-box attacks under different distance metrics, while sacrificing little accuracy on clean images.  Against adaptive full-knowledge attackers creating end-to-end attacks, our system successfully augments the existing robustness of adversarially trained networks, for which our methods are most effectively applied.
\end{abstract}

\section{Introduction}
\label{sec:Introduction}

Modern neural networks provide state-of-the-art classification performance in many applications such as image recognition and language processing.  As such, neural networks are seeing tremendous application growth in every day tasks and thus their safety and security have never been more paramount to human society.  

Despite their increasing sophistication and diminishing error rates, original work by Szegedy et al. \cite{Szegedy} demonstrated that small manipulations to images, called \emph{adversarial examples}, can easily fool neural networks. In a white-box adversarial setting, a malicious actor exploits a knowledge of a neural network's parameters and defenses to intelligently create these examples that are then misclassified by the network. Previous efforts to combat these perturbations rely on several strategies to formulate algorithms or network architectures that are more resistant to adversarial attacks.  The two main camps of defenses either manipulate network learning procedures at training time (adversarial training) \cite{MadryPGD,dingMMA,mustafaRestrictedHiddenSpace,SAT,Trades,featureScatter,TramerEnsemble,NaCascadeTraining} or 
apply transformations to inputs that counter adversarial perturbations before presenting them to the network for classification or detection \cite{PixelDefend,DasSHIELD,ShenAPEGAN,MustafaSuperResolution,XieMitigatingThruRandomization,GuoImageTransformation,EnsembleAbbasi, XuFeatureSqueeze}.  

In response to the previously mentioned defenses, many white-box attack techniques have been developed that have minimized their effectiveness or bypassed them altogether; particularly against the latter camp of defense techniques \cite{CarliniWagner,CarliniMagNetApeGan,HeEnsemblesWeak,ObfuscatedGrads}.  

In this paper, we present a novel system whose elements span the work of both camps.  We adversarially train a neural network and then encapsulate it with custom preprocessing and classification layers.  The core element of our preprocessing technique borrows ideas from neural style transfer \cite{Gatys}: we copy the input once for each possible label and impart the feature level polynomial kernel statistics of sampled training set images from each class onto the copies.  Our classification layer treats the transformed copies as a hard vote committee for classification.  However, by default, the system assumes that the classification on the original input is correct and only changes its decision if a minimum quorum of the committee agrees on a new label. 

We evaluate our defense both \emph{statically} and \emph{adaptively} against a ResNet18 \cite{ResNet} classifying the CIFAR10 \cite{CIFAR10} dataset.

Our system does share similarities to previous defense strategies that have vulnerabilities; namely in the preprocessing and ensembling effects, but we also differ in one fundamental way.  Previous ensembles of preprocessing defenses have been rendered ineffective \cite{HeEnsemblesWeak} by adaptive attackers, but to our knowledge these defenses incorporated processing filters that attempted to `clean' contaminated examples and maintain (or simplify) their salient features in the pixel space.  In contrast, our polynomial kernel optimization strategy splays the input image towards all other labels at the feature level.  Coupled with our classification system, the full-knowledge attacker has the burden of convincing a critical mass of these seemingly opposing transformations to coalesce around a single incorrect label with a single perturbation set or otherwise the perturbed image passes through the baseline adversarially trained network (the default classification unless the committee overrules it).              

We make the following contributions in this paper: 

\begin{itemize}
	\item We construct a novel defense system that shows increased robustness against a diverse set of attacks of varying type, imperceptibility, and strength.
	\item We demonstrate how imparting feature level polynomial kernels from training set sample images to new examples can correctly reclassify weak adversaries and form an ensemble of diversely transformed images.
	\item We form a robust classification system that adapts the final prediction based on both the original input and transformed input copies.
\end{itemize}

\subsection{Preliminary Notations}

A convolutional neural network classifier is considered as a function $f(\vec{x};\theta)$ that maps an input image $\vec{x}\in [0,1]^{C\text{x}W\text{x}H}$ to a vector $\vec{y}\in \R^K$ given $\theta$, a set of network weights and biases.  Element $y_k$ of $\vec{y} = \{y_1, y_2, ... y_K\}$ represents the probability $\vec{x}$ belongs to class $k$ such that $\sum_{k=1}^{K} y_k = 1$ and $K$ is the number of possible distinct labels that can be individually assigned to $\vec{x}$.  A final predicted label, $y_p$, is assigned to $\vec{x}$ by the network using the classification function $Y = \argmax_k(\vec{y})$.  If the ground-truth label of $\vec{x}$, $y_t$, is equal to $y_p$, then the network has correctly classified $\vec{x}$.

Given a training set $T$, a classifier $f(\vec{x};\theta)$ trained on T, and evaluated on a batch of images $X = \{\vec{x}_1, \vec{x}_2, ... \vec{x}_B \}$, an illicit actor may gain unhindered access to $\theta$ and compute a set of perturbations $\Delta = \{\vec{\delta}_1, \vec{\delta}_2, ... \vec{\delta}_B  \}$ to define a set of \emph{adversarial} inputs, $\vec{x}_{adv,b} = \vec{x}_b + \vec{\delta}_b, b=1...B$.  The goal of each adversary is to trick the network into a misclassification $Y(f(\vec{x}_{adv,b};\theta)) \ne y_t|_{\vec{x}_b}$ and more generally the actor maximizes $\E{[\mathds{1}(Y(f(\vec{x}_{adv};\theta)) \ne y_t|_{\vec{x}})]}$.

\subsection{Problem Statement}
\label{sec:ProblemStatements}

In this work, we propose and evaluate defense mechanisms that minimize the effect of adversarial attacks.  Formally,
\begin{enumerate} 
	\label{problems}
	\item Given a trained neural network $f(\vec{x};\theta)$ and set of potentially perturbed images in a white-box adversarial setting, we must present a defense that minimizes $\E{[\mathds{1}(Y(f(\vec{x}_{adv};\theta)) \ne y_t|_{\vec{x}})]}$.
		
\end{enumerate}

\subsection{Threat Model}

The threat model determines under what conditions we claim the defense mechanism to be secure \cite{ObfuscatedGrads, Evaluation}.  We attempt to evaluate performance in a \emph{white-box} adversarial setting with the following assumptions that are specific to our defense:

\begin{enumerate} 
	\label{assumptions}
	\item The attacker knows the neural network's training set, $T$, and the network's weights and biases, $\theta$.
	\item The attacker is aware of the defense mechanism's algorithmic steps, hyperparameter settings, and feature level loss calculations.
\end{enumerate}

Before we present our system we briefly review existing attack methods and previously proposed defenses.

\section{Review}
\label{sec:Review}

First we present attack methods and then the defenses working to mitigate the attacks.

\subsection{Attack Methods}
\label{sec:ReviewofAttackMethods}

\subsubsection{Fast Gradient Sign Method (FGSM)} 

Goodfellow et al. \cite{FGSM} proposed the Fast Gradient Sign Method as an efficient approach to generate adversarial examples based on the gradient of the classification loss function with respect to the image.  An image $\vec{x}$ is perturbed 
to obtain an adversarial image $\vec{x}_{adv}$ as follows:

\begin{equation}
	\vec{x}_{adv} = \vec{x} + \alpha*sign(\nabla_{\vec{x}}L(\vec{x},y_t;\theta)) 
	\label{GoodfellowAdvImage}
\end{equation}
where $\alpha$ is an $L_{\infty}$-bounded step size, $y_t$ is the ground-true label of x, L is a classification loss function such as cross-entropy loss, and $\theta$ is the set of neural network parameters.  FGSM calculates the adversarial example in a single step by taking the sign of the gradient and moving a fixed distance to stay within an $L_{\infty}$-norm boundary around $\vec{x}$.  \\

\subsubsection{Basic Iterative Method (BIM)}

The Basic Iterative Method (BIM) \cite{BIMkurakin} is an iterative version of FGSM using smaller incremental step sizes and clipping pixel values after each update as necessary.  The basic update step to the input is,

\begin{equation}
	\vec{x}_{n+1} = Clip_{\epsilon}\left(\vec{x}_n + \alpha*sign(\nabla_{\vec{x}_n}L(\vec{x}_n,y_t;\theta))\right) 
	\label{InterativeEqnForBIM}
\end{equation}

where $Clip_{\epsilon}$ projects the image to a maximum $L_\infty$-bound set by $\epsilon$. \\

\subsubsection{Projected Gradient Descent (PGD)}

Madry et al. \cite{MadryPGD} generate adversarial examples
from a generalized version of BIM that projects to different $L_p$-bounds.  They have also shown that networks trained using adversarial examples generated by their methods are resilient against adversarial attacks.   

\subsubsection{DeepFool}

Dezfooli et al. \cite{DeepFool} convert the local non-linear problem of finding an adversarial example to that of an affine problem by linearizing the decision boundary between classes and projecting the input image towards a classification boundary in an iterative manner until misclassification occurs.  This process extends to the multinomial classification problem by approximating the non-linear boundaries of multiple classes surrounding $\vec{x}$ as a polyhedron and finding the minimum distance to its surface.\\

\subsubsection{Carlini-Wagner}

Carlini et al. \cite{CarliniWagner} formulate an optimization problem that minimizes the $L_{p}$-norm distance between the clean and adversarial image, subject to the constraint that the adversarial image is misclassified by the neural network.  They move the misclassification constraint into the objective function to form the non-convex problem:

\begin{equation}
	\begin{aligned}
		\min \quad & || \vec{x}_{adv} - \vec{x} ||_p + c*f(\vec{x}_{adv})\\
		\textrm{s.t.} \quad & \vec{x}_{adv} \in [0,1]^n 
	\end{aligned}
\end{equation}
\noindent
where $f(\vec{x}_{adv})$ is a function that signals when the adversary is found and $c$ is a hyperparameter that controls the relative objective weights of minimizing distance and finding the adversary.  The authors propose several different functions for $f$ and show general trends for selecting $c$.  We make special note that a confidence parameter, $\kappa$, may be prescribed to the increase the confidence of misclassification at the expense of higher distortion.   Carlini-Wagner (CW) attacks are strong and unbounded attacks that seek imperceptibility and are employed in our evaluations.  \\

\subsubsection{Backward Pass Differentiable Approximation and Expectation Over Transformation}

Although not technically new attacks, backward pass differentiable approximation (BPDA) and expectation over transformation (EOT) \cite{ObfuscatedGrads, AthalyeSynthesis} allow existing attacks to handle defense systems that apply non-differentiable or randomized techniques.

Athalye et al. point out that these defenses effectively corrupt available gradient information through \emph{shattering} (making gradients unavailable at the input due to a non-differentiable operation), \emph{stochasticity} (applying randomness such that a single calculation of gradients is not representative), and \emph{explosion/vanishment} (artificial overflow or underflow of gradient values).  From a high level, by sampling transformations in the forward pass (EOT) and approximating non-differentiable operations as the identity function in the backward pass (BPDA), gradient based attacks can still find approximate gradients at the input and over many iterations still find a successful perturbation set.

\subsubsection{Decision-Based Attack}

In situations where gradient information is not reliable, a decision-based attack, like the Boundary Attack proposed by Brendel et al. \cite{brendelDecision}, relies only on the forward pass prediction of the network.  The Boundary Attack finds an initial perturbation distribution that is adversarial to the network and then progressively draws new perturbations that both maintain misclassification and reduce the vector norm distance to the original image.

\subsection{Review of previously proposed defenses}
\label{sec:ReviewOfProposedDefenses}

\subsubsection{Adversarial Training}

Conceptually, the adversarial training approach forces the network to learn more robust decision boundaries at training time via exposure to training examples generated using one or more adversarial attack methods or modifications to the learning process.

Madry et. al \cite{MadryPGD} increased network robustness by perturbing training examples up to a fixed $L_p$-bound using PGD at each epoch and letting the network learn from the adversarial distribution.  As an extension of \cite{MadryPGD}, Ding et. al \cite{dingMMA} proposed in Max-Margin training (MMA) that the $L_p$-bound placed on adversarially generated examples need not be fixed and in fact could be optimized on a per example basis to optimize the average decision boundary margin.  Zhang et. al \cite{Trades} included a regularization term in their training loss that searches for an adversarial example in an $\epsilon$-neighborhood that maximally changes the output of the network compared to the clean example before updating the network parameters, thus encouraging decision boundaries to move maximally away from the data.

Further advancement in adversarial training focused on crafting adversarial examples based on not only classification loss, but also on the inter-example feature relationships between clean and generated examples.  In \cite{featureScatter}, a PGD adversarial distribution is selected based on maximizing an optimal transport distance from the clean example distribution at the feature map level resulting in adversarial examples that are diverse and distant from clean examples during training. 

\subsubsection{Image Transformation and Reconstruction}

A different defensive strategy is to apply voluntary \emph{transformations} to inputs or network behaviors at inference time in the hope that these effects cannot be easily approximated by the attacker.  Examples of image transformation include random resizing and padding \cite{XieMitigatingThruRandomization}, JPEG compression \cite{DasSHIELD}, and image quilting \cite{GuoImageTransformation}.  

A related defensive strategy involves attempting to eliminate adversarial perturbations through various means \cite{DasSHIELD,MustafaSuperResolution,PixelDefend, ShenAPEGAN}.  The basis for this strategy relies on the observation that clean and adversarial images, despite often visually similar in pixel space, come from different manifolds in feature space and that any mapping which bridges these two distributions could reduce adversarial corruptions in an image.  

In the next section, we present the polynomial kernel matrix, which we use to transform the feature maps of denoised input images. 

\section{Polynomial Kernels for Image Transformation}
\label{sec:NotationsUsed}

A convolutional neural network, $f(\vec{x};\theta)$, can be viewed as a composition of functions:

\begin{equation}
	f(\vec{x} ; \vec{\theta}) = (F_S \circ F_{fc} \circ F_L  \circ F_{L-1} \circ ... \circ F_1)(\vec{x})
	\label{eqn:DNNasAfunction}
\end{equation}

where $F_S$ is the softmax activation function, $F_{fc}$ is a sequence of fully connected layers, and $F_{1} ... F_{L}$ are convolutional layers, that perform the \emph{feature extraction} function of the network.  Each $F_{l}$, $l \in \{1...L\}$, is itself a composition of functions, transforming the output of the convolutional layers before it,

\begin{equation}
	F_l(\vec{x};\theta_{l}) = [h \circ [Conv(F_{l-1} \circ ... \circ F_1)]](\vec{x})
	\label{eqn:OneConvLayer}
\end{equation}

such that $h$ is a non-linear mapping and $Conv(\cdot)$ is a convolutional operation.  

Let the set $V_l = \{v_1,v_2,...v_C\}$ represent the C feature maps of dimension $H$x$W$ at layer $l$.

Given any pair of feature maps within the same layer, $v_i$ and $v_j$, we define a polynomial kernel function,

\begin{equation}
	\kappa(v_i,v_j) = \left(\left<v_i,v_j\right> + e\right)^d 
	\label{eqn:kernelMatrix}
\end{equation}
where $e\ge0$ and integer $d\ge1$ are chosen parameters and $\left<\cdot,\cdot \right>$ is the inner product.   Treating input maps $v_i$ and $v_j$ as having dimension $N=H \times W$, the kernel function $\kappa(\cdot,\cdot)$ 
has the property that there exists a mapping
$$\phi(v_i): \R^N \rightarrow \R^{\binom{N+d}{d}}$$ such that,  
\begin{equation}
	\kappa(v_i,v_j) = \left<\phi(v_i),\phi(v_j)\right> 
	\label{eqn:PropetyOfkernelFn}
\end{equation}

Given a set of feature vectors $V$ in a convolutional layer, the elements of a kernel matrix $G \in S_+^{C\text{x}C}$ for that layer are computed by,

\begin{equation}
	G_{i,j} = \kappa(v_i,v_j)
	\label{eqn:DefnKernelMatrix}
\end{equation}

The kernel matrix is a positive semi-definite matrix unique to each layer. Note that for $e= 0$ and $d=1$, the kernel matrix is identical to the Gram matrix.

Specifically for polynomial kernels, the mapping $\phi(v_i)$ transforms the input feature map into a higher dimensional vector whose components are formed by all monomial combinations of the input vector components up to total order d.  For example, if $N=2, m=1$, and $d=3$, then $\phi(v_i): \R^2 \rightarrow \R^{10}$ maps the components of $v_i = [v_{i1},v_{i2}]$ to a vector that includes all monomial terms from the set $\{a_{r,s}v_{i1}^rv_{i2}^s\ |\ r,s = 0...3,\ r+s \le 3, a_{r,s} \in \R\}$.

The next section details our defense system construction and the conceptual ideas behind our design choices.

\section{Proposed Training and Defense Methods}
\label{sec:ProposedFramework}

\begin{figure} [ht]
	\centering
	\includegraphics[width=0.5\linewidth]{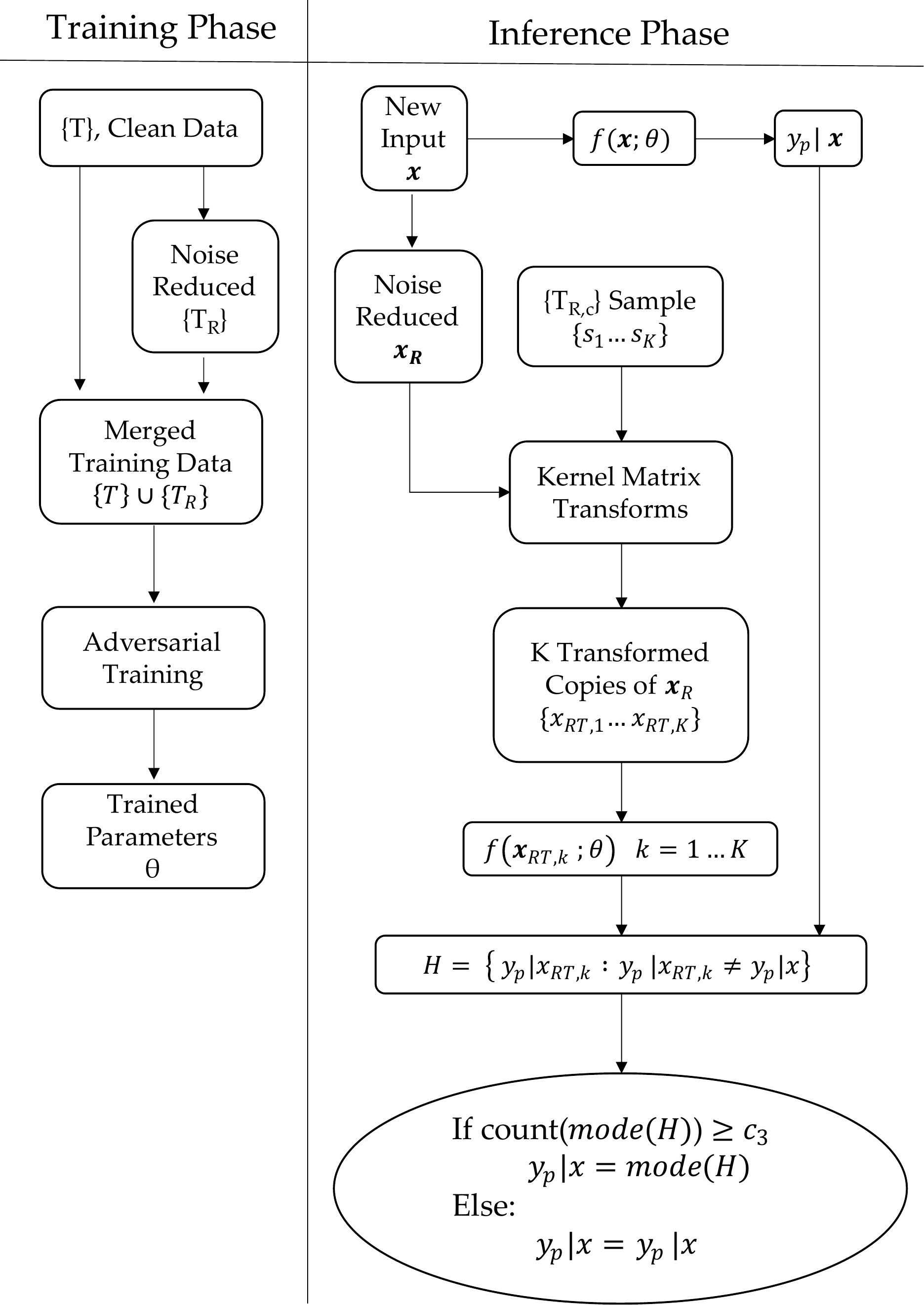}
	\caption{Schematic of proposed training and defense processes.}
	\label{fig:defenseschematic}
\end{figure}

As shown in Fig. \ref{fig:defenseschematic}, we describe a training phase (left column) and defense methodology (right column) to produce a perturbation resistant deep neural network.  All of the steps contained in the following section can be found in Fig. \ref{fig:defenseschematic}.

\subsection{Training Phase}

\emph{Train neural network on clean and smoothed data.}  Since our defense system applies a traditional image smoothing operation such as a median filter to the image, we wish for the network to be able to recognize the smoothed versions of the original image.  Therefore, for each image $t \in T$, we generate a corresponding smoothed image $t_R$ to obtain a set $T_R$ and then use all examples in $\{T\} \cup \{T_R\}$ to train the network.

\subsection{Inference Phase}

\emph{Receive new input $\vec{x}$.}  Classify $\vec{x}$ with $f(\vec{x};\theta)$ and store result $y_p|\vec{x}$.

\emph{Apply smoothing operation to $\vec{x}$ to create $\vec{x}_R$.}  Apply the same smoothing operation to $\vec{x}$ that was used to generate set $\{T_R\}$ in the training phase.

\emph{From the correctly classified, denoised training data set $\{T_{R,c}\}$, draw one sample image corresponding to each class to create a sample set $\{\vec{s}_1...\vec{s}_K\}$.  Compute and store the polynomial kernel matrices for each sample image.}  The $K$ stored polynomial kernel matrix sets from $\{\vec{s}_1...\vec{s}_K\}$ will act as target kernel matrices towards which our smoothed $\vec{x}_R$ will be transformed.

\emph{Create $K$ copies of $\vec{x}_R$.}  Each $\vec{x}_R$ copy's polynomial kernels will be transformed independently towards one of the $K$ sets of polynomial kernels stored from $\{\vec{s}_1...\vec{s}_K\}$ in the next step.

\subsubsection{Iteratively impart polynomial kernel values of $\vec{s_k}$ to $\vec{x_{R,k}}$.}
\begin{figure} [ht]
	\centering
	\includegraphics[width=0.6\linewidth]{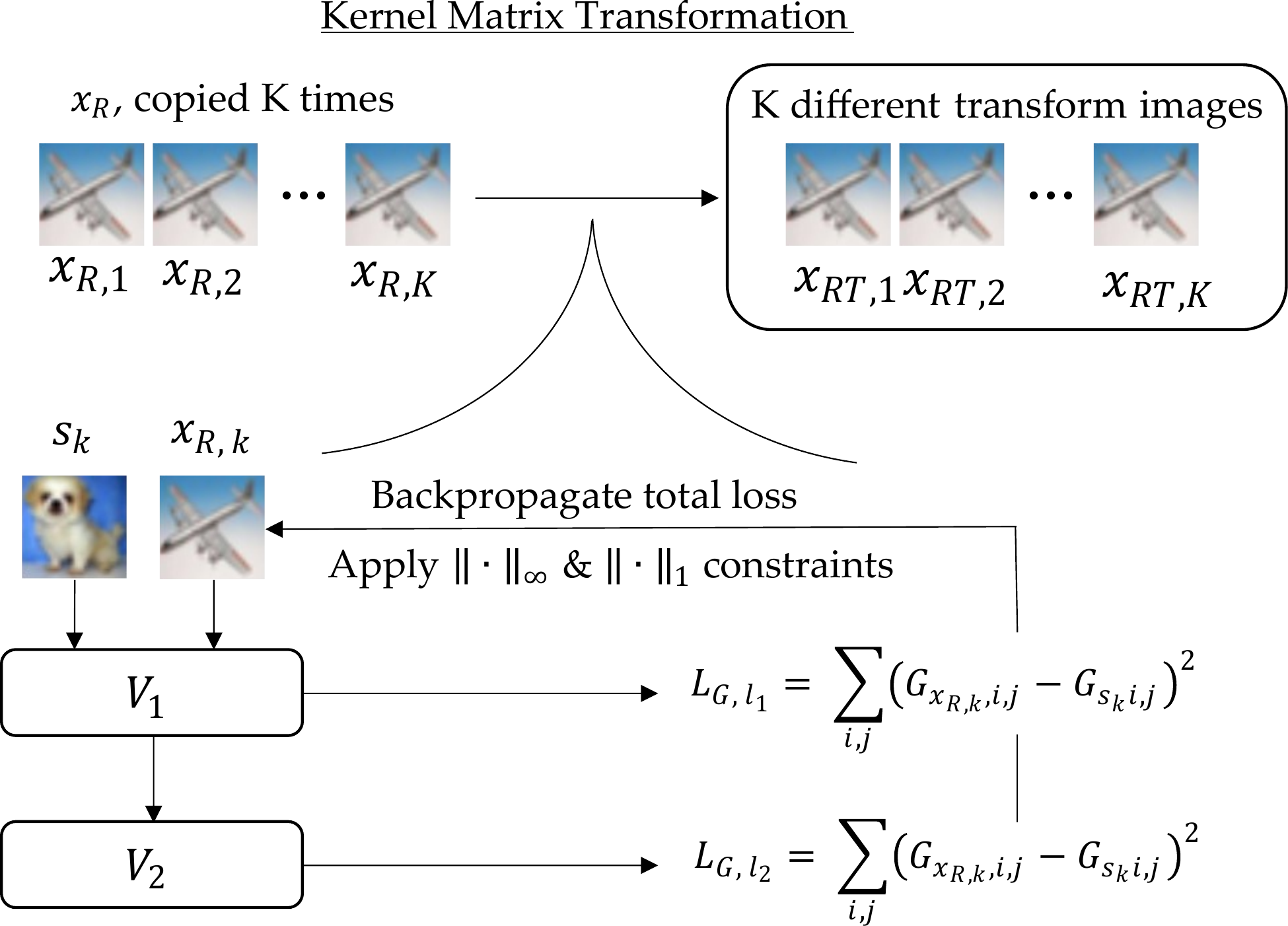}
	\caption{Schematic of polynomial kernel transformation procedure.}
	\label{fig:transformschematic}
\end{figure}

%\textbf{Iteratively impart polynomial kernel values of $\vec{s_k}$ to $\vec{x_{R,k}}$.}  
The following discussion is based on the flowchart in Fig. \ref{fig:transformschematic}.  Conceptually similar to but substantially different from neural style transfer \cite{Gatys}, the polynomial kernel module computes losses in one or more layers of the network comparing the kernel matrix values of the $k^{th}$ copy of $\vec{x}_R$ to the $k^{th}$ sampled image $\vec{s}_k$ (from class $k$).  The MSE loss for a single layer is:
%and ($\vec{x_{R,k}}$,$\vec{s_k}$) pair is:

\begin{equation}
	L_{G,l} = \sum_{i,j} \left({G_{\vec{x}_{R,k},i,j} - G_{\vec{s}_{k},i,j}}\right)^2
	\label{eqn:KernelLossCalculation}
\end{equation}

We compute MSE losses at chosen network feature extraction layers and backpropagate their summed loss to the input, updating $\vec{x}_{R,k}$ using a gradient descent algorithm.  After every pixel update, we apply both an $L_1$ \cite{DuchiLOne} and $L_{\infty}$ distance constraint to the current updated input, limiting the vector distance between $\vec{x}_{R,k}$ and its final transformed version $\vec{x}_{RT,k}$.  The values of these constraints are hyperparameters obtained from validation study.  We begin the iteration process by randomly perturbing the target $\vec{x}_R$ within the chosen $L_{\infty}$ ball.

We iterate on the loss-backpropagate-constraint loop for a chosen number of iterations, recalculating the layer losses, pixel gradients, and vector norm projections independently for each iteration.  We emphasize that this transfer procedure is completed for each $\vec{x}_{R,k}$ corresponding to a \emph{different} polynomial kernel set generated from $\vec{s}_k$.  The end result is that we output $K$ differently transformed versions of the input, $\vec{x}_{RT,1} ... \vec{x}_{RT,K}$, to the next step.

\subsubsection{Form voting committee of transformed image predictions and classify $\vec{x}$ based on images that changed label prediction from $f(\vec{x};\theta)$.}

Having transformed all $K$ copies of $\vec{x}_R$, we run inference on each transformed copy through the network and store each label prediction in a set, $\{y_p|\vec{x}_{RT,1} ... y_p|\vec{x}_{RT,K}\}$.  From this set of predictions, we create a subset $H$ keeping only the predictions that are \emph{different} than the stored, original prediction on $\vec{x}$.  We then apply a simple hard vote type rule to the remaining predictions: if the most common predicted label, $mode(H)$, of the predictions that changed due to transformation has a count greater than or equal to a threshold hyperparameter $c_3$, we change our final prediction to this most common label.  Otherwise, we keep the original network prediction on $\vec{x}$.

\subsection{Defense System Intuition}
\label{sec:DefenseSystemIntuition}

\subsubsection{From an Attacker's Perspective}
\begin{figure}
	\centering
	\includegraphics[width=0.4\linewidth]{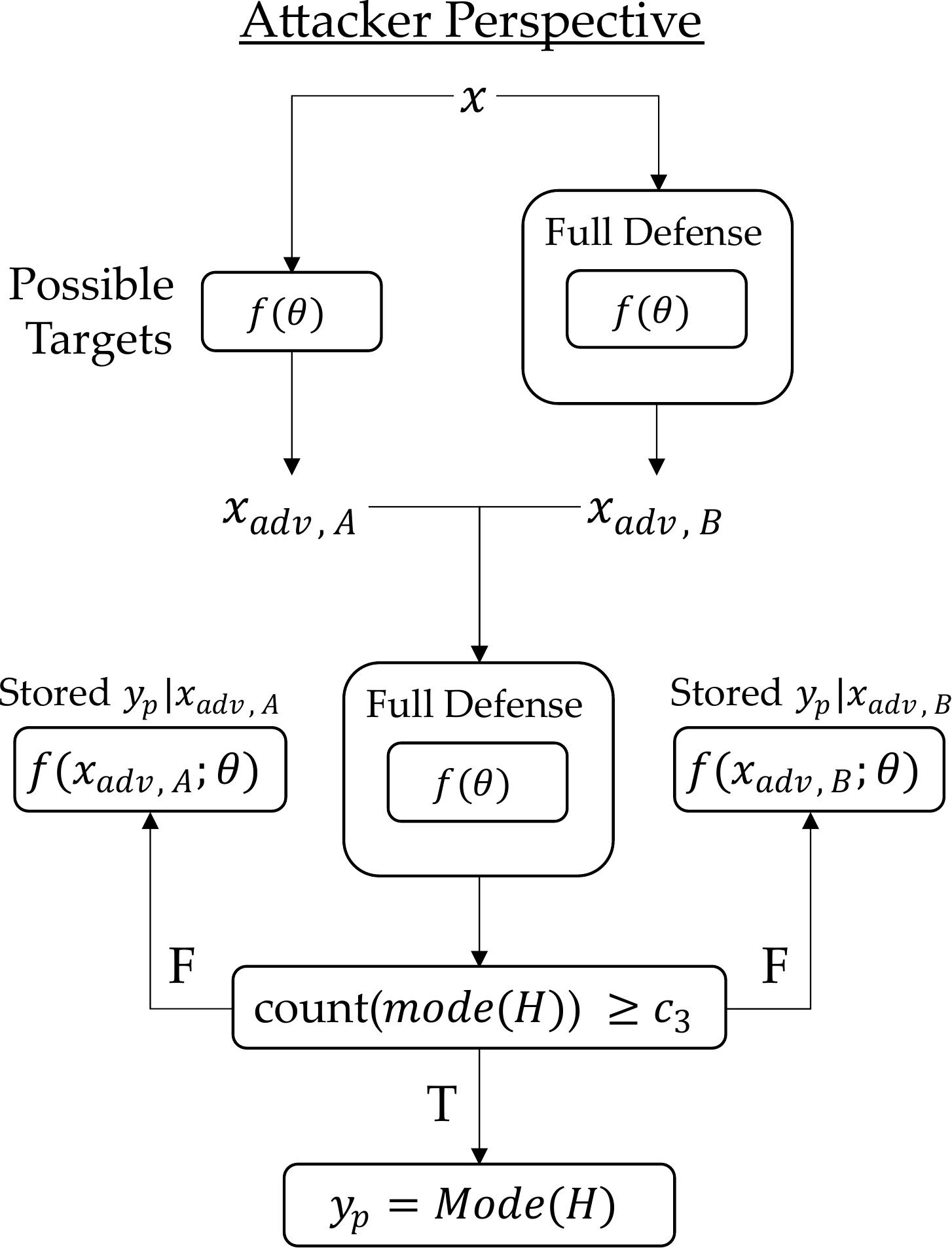}
	\caption{Decision flowchart available to attacker showing two possible scenarios.}
	\label{fig:attackerperspective}
\end{figure}

As shown in Fig. \ref{fig:attackerperspective}, the attacker can choose to attack the network directly (option A) or simulate the full defense system (option B) by implementing the defense algorithm on the forward pass, using gradient approximations on the backward pass, and increasing their sample sizes to better estimate the polynomial kernel transformations that the defender will use at inference time.  These two different choices would result in one of two different adversarial images, $\vec{x}_{adv,A}$ or $\vec{x}_{adv,B}$, being fed to the defense system.  The defense system initializes by making a prediction on the adversarial image through the `bare' network, thus storing either $f(\vec{x}_{adv,A};\theta)$ or $f(\vec{x}_{adv,B};\theta)$.  Note how $\vec{x}_{adv,B}$ was \emph{not} designed to attack the bare network $f(\theta)$ without the defense transformations.  Adversarial image $\vec{x}_{adv,B}$ could still transfer to $f(\theta)$, but we posit that it cannot be more likely to succeed than if the attack algorithm had just attacked $f(\theta)$ directly, i.e. crafted $\vec{x}_{adv,A}$ in the first place.

The ideal scenario B outcome for the attacker is that a multitude (greater than threshold parameter $c_3$) of the polynomial kernel transforms move $\vec{x}_{adv,B}$ to the same incorrect label; otherwise the network effectively drops its own defense and resorts to $f(\vec{x}_{adv,B};\theta)$, for which we have already stated $\vec{x}_{adv,B}$ was not designed.  

\subsubsection{The Polynomial Kernel Transforms}

\begin{figure*} [!ht]
	\includegraphics[width=\textwidth]{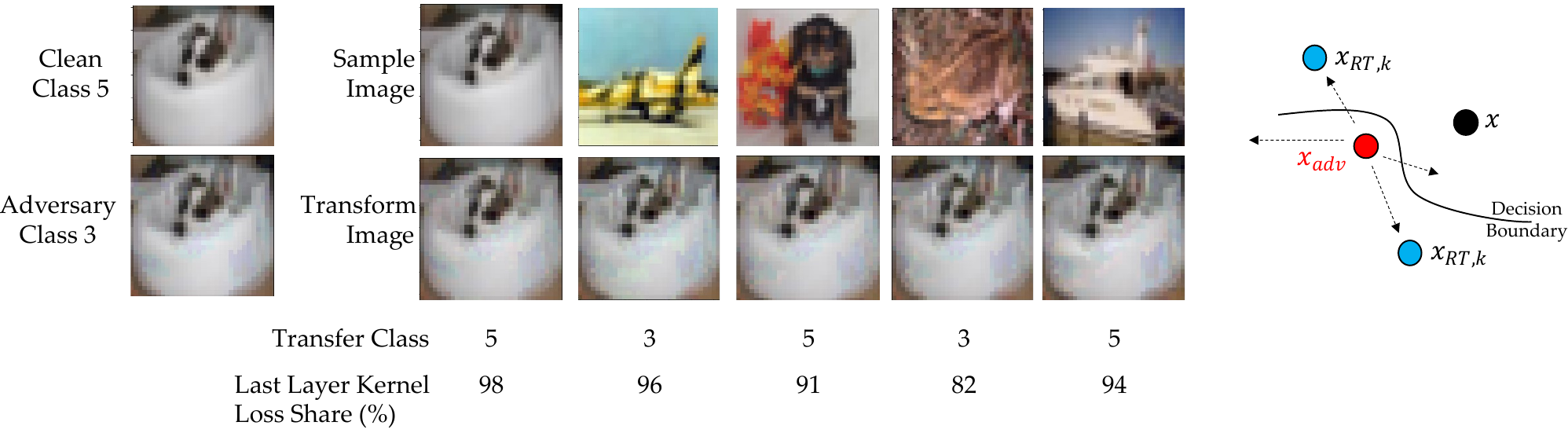}
	\caption{Simple toy example of polynomial kernel transformation for a CIFAR10 'dog' image.  Adversary created using 25 iterations of Basic Iterative Method ($\epsilon=8/255$) against an adversarially trained ResNet18 network.  Each 'sample' image's polynomial kernel values are imparted onto the adversary to create a separate 'transform' image.  Some of the pixel variations between the clean, adversarial, and transformed examples can be seen on the dog's bed surface.}
	\label{fig:ImageTransferExample}
\end{figure*}

Recall our defense system takes $K$ (number of classes) copies of the smoothed input $\vec{x}_R$ and iteratively imparts the polynomial kernel values of $K$ sampled images from the clean training set (one sample from each class).  In the toy example of Fig. \ref{fig:ImageTransferExample}, a CIFAR10 'dog' is initially correctly classified as class 5, but incorrectly classified to class 3 'cat' when the ResNet18 network is directly attacked (BIM-scenario A).  We draw a sample of images from a few classes and also include the original clean image, then apply 25 iterations of kernel transformation from the samples to the adversarial example using feature maps from each residual block in ResNet18 (early and deep layers).  Several transformations are successful in restoring the correct classification and some are not.

Our results suggest several important conclusions.  First, if the kernel matrix information of the original image is imparted on a perturbed version of itself, the network recovers the correct classification.  Second, transformations derived from a sample class $k'$ can recover the correct classification of images from other classes $k \ne k'$ as can be seen from the boat sample reclassifying the dog.  Third, we track the average polynomial kernel loss \ref{eqn:KernelLossCalculation} at iteration 0 for the deepest residual block in ResNet18 as a $(\%)$ share of the total loss across all layers and recognize the original image `sample' has the highest loss share compared to other samples; likewise, the early layers have relatively low kernel loss.  This effect implies that adversarial attacks' imperceptible pixel manipulations create correspondingly small changes in early layer kernel matrices, but larger changes in the deeper layers, which are closer to classification.  Lastly, not all transformations result in correct classifications.  An important consideration not shown in this particular example, is sometimes we receive random classifications $k'' \notin \{k,k'\}$.

Our interpretation of the toy experiment is pictured on the far right of Fig. \ref{fig:ImageTransferExample}, a clean example $\vec{x}$ (black) is perturbed across the decision boundary to a misclassified $\vec{x}_{adv}$.  Several transformation directions are shown altering $\vec{x}_{adv}$ to $\vec{x}_{RT,k}$ depending on the sample's class.  Some transformations cross back over the decision boundary and are correctly classified, some remain misclassified, and others end up in random other classes.  Let the tuple $(k',k'')$ represent the decision boundary between class $k'$ and $k''$.  We believe and have found that if the adversary is close to the boundary of the correct class or 'weak', then $(k,k')$ presents as a larger cross-sectional target to the kernel transforms as compared to others in $\{(k'',k), (k'',k'),(k'',k'')\}$.  Therefore, if any reclassification is to occur, a propensity exists for a correct classification over a random classification.  What about clean examples, will they be mistakenly reclassified?  Particularly in the case of adversarially trained networks, the clean data is pushed away from the boundary, so the risk of any reclassification is lowered compared to a weak adversary pushed just over the boundary.  Further, we perform validation studies to choose a voting threshold parameter $c_3$ that allows for a few random reclassifications of clean examples without the defense changing its prediction due to the voting committee. 

How do these ideas interact with the attack scenario of Fig. \ref{fig:attackerperspective}?  The polynomial kernel transforms are not attempting to cleanse adversarial perturbations, rather they are independent feature transformations to all possible classes.  The scenario B attacker, to achieve the ideal scenario, must coalesce these different transformations with a single perturbation set or risk that the designed adversary is transferred back to the bare network as a weaker adversary than if it had just targeted the bare network originally.   

%[ht]

\section{Empirical Evaluation}
\label{sec:EmpricalEvaliation}

\subsubsection{Evaluation networks and datasets}
\label{sec:EvaluationNetworksAndDatasets}

To evaluate our defense system, we applied our techniques to a basic Resnet18 neural network \cite{ResNet} and tested it against a variety of attacks on the CIFAR10 dataset \cite{CIFAR10} in the Pytorch environment \cite{pytorch} with 1 GPU.  CIFAR10 examples consist of 50,000 training and 10,000 test 32x32 3-channel color images of various natural objects ranging from airplanes to dogs. We trained the network in two different ways.  In the first 'Std' training, the network was trained for 100 epochs using standard cross-entropy loss and in the second 'Adv' training, the network was adversarially trained for 100 epochs using the free adversarial training method (m=5) \cite{shafahiFree} and TRADES loss \cite{Trades} with $\lambda=1$.

\subsubsection{Hyperparameter values}
\label{sec:DeterminationOfHyperparameterValuesAndFilterLocations}

We segmented off 400 randomly selected points from the CIFAR10 test set and performed validation studies with the trained networks to find suitable settings for kernel layer selections, $L_1$-limits ($c_1$), $L_\infty$-limits ($c_2$), voting thresholds $c_3$, transform iterations, initial learning rate, and optimizer momentum.  We employed the RMSprop \cite{RMSprop} optimizer ($\rho=0.1$, $lr=0.07$) for updating input images.  The kernel layers are indexed from [0,1,2,3,4,5], referring to `after the 1st convolution', after `the 1st ReLU()', `after residual block 1', `after residual block 2', and so on in ResNet18.  Our smoothing operation was a median filter with window size 2x2.

\subsubsection{Generation of Adversarial Examples}
\label{sec:GenarationOfAversarialExamples}

We used the FoolBox \cite{Foolbox1,FoolBox2} library to create a set of $L_\infty$-bounded (BIM) and unbounded attacks (CW-L2, DeepFool) generated against 1000 randomly selected CIFAR10 test set images (drawn separately from the validation study points).  The Carlini-Wagner attacks optimized the $L2$ distance metric and executed 15 binary search steps beginning with $c=0.01$.  For scenario B adaptive attacks, we used BPDA across the kernel transforms and Pytorch subgradients across the median filter.  To handle the ensemble function, we modified the logit output similar to \cite{HeEnsemblesWeak}.  If $Z_k$ is the output logit vector of sample $s_k$, then our adaptive attack model outputs $\sum_{k=1}^{K}{\frac{Z_k}{||Z_k||}}$.   

\begin{table}
	\centering
	\caption{Defense System Hyperparameters}
	\label{params}
	\begin{tabular}{lcc}\toprule
		Parameter & Std & Adv \\\midrule
		$c_1$ & 30.0 & 30.0  \\
		$c_2$ & .02 & .02  \\
		$c_3$ & 3 & 9  \\
		$Iterations$  & 10 & 10 \\\bottomrule
	\end{tabular}
\end{table}

\subsection{CIFAR10 Evaluation}
\label{sec:BenchmarkDataset2}

Our results in Table \ref{cifarresults2} show several important trends.  We applied our defense system, although intended for an adversarially trained network, to a standard trained network for completeness.  Those results show that while the system correctly classifies static scenario A adversarial examples quite frequently, particularly the weaker $L_2$ adversaries (DeepFool, CW $\kappa=0$), the system is completely bypassed in scenario B by stronger attacks.  

In the case of an adversarially trained network under an adaptive BIM $\epsilon=8/255$ attack (scenario B), the defense system maintains a minimum accuracy of $40.4\%$ compared to the baseline network accuracy of $38.6\%$ when we use polynomial kernel transforms from layers 1-5.  It does \emph{not} maintain this accuracy and in fact is reduced from the baseline when we use kernel matrices from only early layers in the network (0-2).  This difference in performance suggests that our system must impart information from the deeper feature extraction layers, which contain more class-specific features.

We notice a bifurcation in trend between the weaker (DeepFool, CW $\kappa=0$) and stronger (CW $\kappa=5$) attacks between scenarios A and B, where the system improves accuracy against weaker attacks, but lowers accuracy against stronger attacks.  As stated before, we believe the system is better suited for weaker attacks and thus in this case, these attacks close to the decision boundary are failing to find a consensus incorrect label in the ensemble and transferring the perturbed image back to the bare network for classification.  The system is not as adept at handling stronger attacks and presumably the attacks can take advantage of our transformations by applying small perturbations in the image space that are favorably magnified by our transforms, so we see the opposite trend against CW $\kappa=5$ attacks.  Increasing $\kappa$ has also been shown to increase the transferability of an attack \cite{CarliniWagner} and would more likely cause our bare network to misclassify even if a voting threshold is not reached.  Appropriately, if we were to increase $\kappa$ further, the system would collapse, but at increasing risk to the attacker that perturbations becomes visually perceptible.  Lastly, using a higher order polynomial kernel (d=2) did not provide a benefit in performance and in most cases hurt the system.

\begin{table} [ht]
	\centering
	\caption{CIFAR10 Test Set Results - ResNet18}
	\label{cifarresults2}
	\begin{tabular}{lccccccccc}\toprule
		Network \& Defense & Training & Scenario & Layers & Clean & DeepFool & CW & CW  & BIM & BIM \\\midrule
		Parameters: & - & - & - & - & - & $\kappa = 0$ & $\kappa = 5$ & $\epsilon$ = 4 & $\epsilon$ = 8 \\ [.1cm]
		Attack Iterations: & - & - & - & - & 50 & 20 & 20 & 20 & 20 \\ [.1cm]
		ResNet18, No Defense & Std & A & - & 95.8  & 0.0  & 0.0  & 0.0  & 0.0  & 0.0  \\ [.1cm]
		ResNet18, Full $e,d = 0,1$ & Std & A & 0-2 & 92.8  & 92.7  & 93.0  & 88.5  & 32.5  & 4.7  \\ [.1cm]
		ResNet18, Full $e,d = 0,1$ & Std & B & 0-2 & 92.8  & 24.9 & 27.2  & 0.0  & 0.0  & 0.0 \\\midrule
		ResNet18, No Defense & Adv & A & - & 87.0 & 0.0  & 8.0  & 64.0  & 64.5  & 38.6   \\[.1cm]
		ResNet18, Full $e,d = 0,1$ & Adv & A & 0-2 & 84.7 & 56.5 & 63.9 & 61.3  & 66.2 & 43.1  \\[.1cm]
		ResNet18, Full $e,d = 0,1$ & Adv & A & 1-5 & 84.6 & 46.1 & 51.6 & 61.2 & 65.3 & 41.4  \\[.1cm]
		ResNet18, Full $e,d = 1,2$ & Adv & A & 1-5 & 84.5 & 49.1 & 54.6 & 61.2 & 65.2 & 42.0 \\\midrule
		ResNet18, Full $e,d = 0,1$ & Adv & B & 0-2 & 84.7 & 75.4 & 76.2 & 48.5 & 62.8 & 38.4  \\[.1cm]
		ResNet18, Full $e,d = 0,1$ & Adv & B & 1-5 & 84.6 & 72.9 & 77.8 & 46.9 & 66.4 & 40.4 \\[.1cm]
		ResNet18, Full $e,d = 1,2$ & Adv & B & 1-5 & 84.5 & 62.5 & 74.3 & 47.1 & 65.2  &  39.6 \\\bottomrule
	\end{tabular}
	\stoptable{Results in \%.  Layers column indicates chosen feature extraction layers to impart kernel transforms.}
	
\end{table}

\section{Conclusion}
In this paper, we developed an adversarial defense system that improves the robustness of a neural network against a diverse array of white-box attacks.  Our GPU resources limited our generated attacks to a small number of iterations compared to previous works \cite{CarliniWagner}, but on a comparative basis our defense maintained increases in robust accuracy against \emph{adaptive} attacks under certain hyperparameter settings.

\bibliographystyle{unsrt}
\bibliography{citations}

\end{document}